\IfFileExists{ieeeconf.cls}{%
\documentclass[letterpaper,10pt,conference]{ieeeconf}
}{%
\documentclass[conference,10pt,letterpaper]{IEEEtran}
}

\IEEEoverridecommandlockouts
\providecommand{\overrideIEEEmargins}{}
\usepackage{amsmath,amssymb,amsfonts}
\usepackage{graphicx}
\usepackage{booktabs}
\usepackage{multirow}
\usepackage{cite}
\usepackage{xcolor}
\usepackage{url}
\usepackage[hidelinks]{hyperref}
\usepackage{pifont}                 
\let\labelindent\relax             
\usepackage{enumitem}
\usepackage{xspace}

\makeatletter
\let\original@makecaption\@makecaption
\long\def\@makecaption#1#2{%
  \ifx\@captype\@IEEEtablestring
    {\footnotesize\normalfont\raggedright\noindent #1: #2\par}%
    \vskip 6pt
  \else
    \original@makecaption{#1}{#2}%
  \fi
}
\makeatother

\makeatletter
\renewcommand{\footnoterule}{\kern-3\p@\hrule\@width0.4\columnwidth\kern2.6\p@}
\makeatother

\makeatletter
\providecommand{\bstctlcite}{\@ifnextchar[{\@bstctlcite}{\@bstctlcite[@auxout]}}
\def\@bstctlcite[#1]#2{\@bsphack
  \@for\@citeb:=#2\do{%
    \edef\@citeb{\expandafter\@firstofone\@citeb}%
    \if@filesw\immediate\write\csname #1\endcsname{\string\citation{\@citeb}}\fi}%
  \@esphack}
\makeatother

\definecolor{darkblue}{HTML}{234A6F}

\newcommand{\noindentp}[1]{\par\vspace{0.1em}\noindent\textbf{#1}}
\newcommand{\secv}{\vspace{-0.1em}}
\newcommand{\ssecv}{\vspace{-0.1em}}
\newcommand{\pos}{\mathbf{p}}                 
\newcommand{\rot}{\mathbf{R}}                 
\newcommand{\pose}{\mathbf{T}}                
\newcommand{\relpose}{\tilde{\mathbf{T}}}     
\newcommand{\force}{\mathbf{f}}               
\newcommand{\wrench}{\mathbf{w}}              
\newcommand{\fcmd}{\force^{\mathrm{cmd}}}     
\newcommand{\fmeas}{\force^{\mathrm{meas}}}   
\newcommand{\fext}{\force^{\mathrm{ext}}}     

\newcommand{\wmeas}{\wrench^{\mathrm{meas}}}
\newcommand{\pihl}{\pi_{\theta}}              
\newcommand{\obs}{\mathbf{o}}                 
\newcommand{\act}{\mathbf{a}}                 
\newcommand{\state}{\mathbf{s}}               
\newcommand{\img}{\mathbf{I}}                 
\newcommand{\integ}{\mathbf{I}}               
\newcommand{\dx}{\Delta\mathbf{x}}            
\newcommand{\cphase}{c}                       
\newcommand{\hands}{\mathcal{H}}              
\newcommand{\keypts}{\mathcal{K}}             
\newcommand{\horizon}{H}                      
\newcommand{\norm}[1]{\left\lVert #1 \right\rVert}

\providecommand{\noindentp}[1]{\vspace{2pt}\noindent\textbf{#1}~}   
\providecommand{\cmark}{\textcolor{green!60!black}{\ding{51}}}        
\providecommand{\xmark}{\textcolor{red}{\ding{55}}}                  
\providecommand{\task}[1]{\textsc{#1}}                               
\providecommand{\bl}[1]{\texttt{#1}}                                 
\providecommand{\ours}{\texttt{Ours}\xspace}
\providecommand{\figorbox}[3]{\IfFileExists{#1}{\includegraphics[width=#2]{#1}}{\fbox{\parbox[c][#3][c]{\dimexpr#2-2\fboxsep-2\fboxrule\relax}{\centering\small placeholder:\\ \detokenize{#1}}}}}

\usepackage{algorithm}
\usepackage{algpseudocode}
\newcommand{\modelname}{\textsc{FLoRe}\xspace}
\makeatletter
\newcommand{\teaserfigure}{%
\begin{minipage}{\textwidth}
\centering
\includegraphics[width=\textwidth]{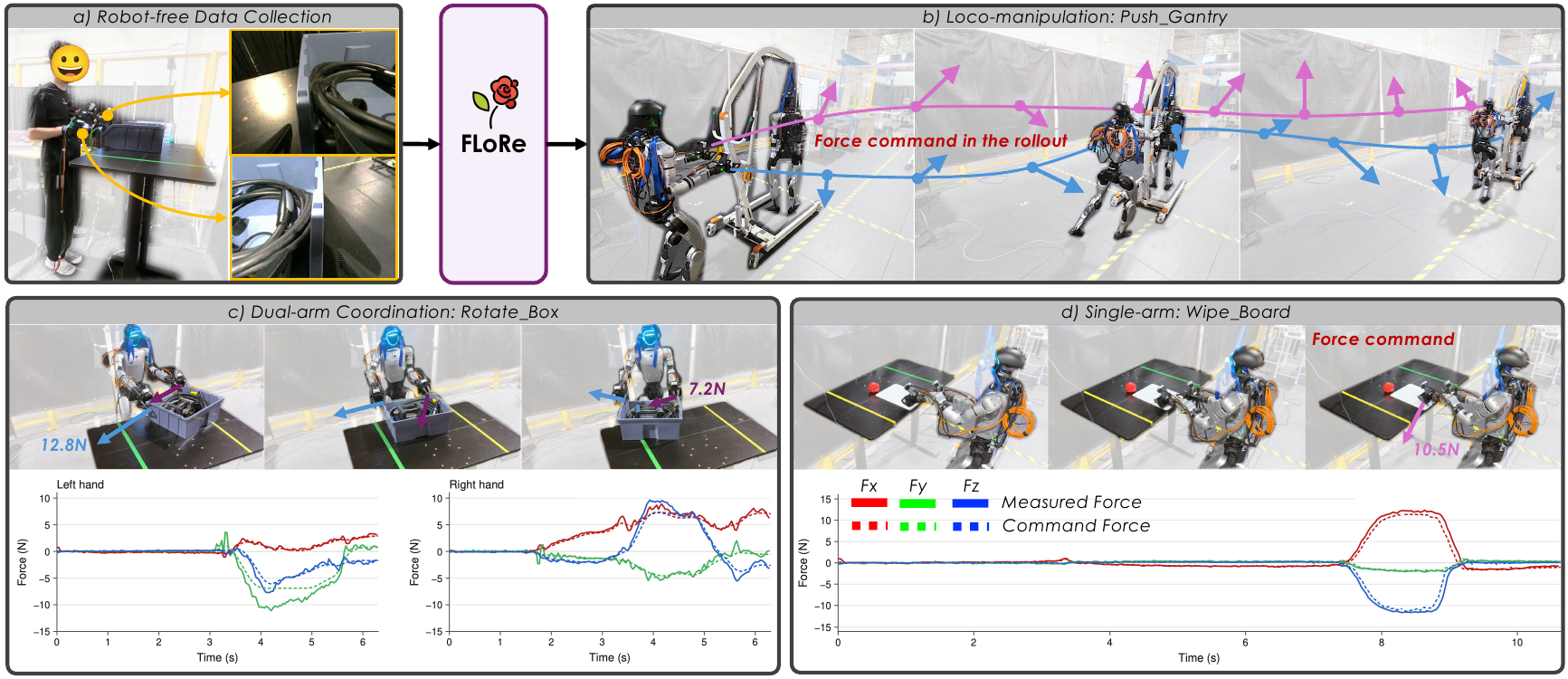}%
\setcounter{figure}{0}%
\def\@captype{figure}%
\caption{
\textbf{\modelname is a framework that learns force-reactive loco-manipulation from robot-free demonstrations.}
\textbf{(a)}~Two handheld UMI-style grippers carrying 6-axis F/T sensors and three trackers record hand/torso motion, gripper width and contact forces.
From these demonstrations, we train a force-reactive diffusion policy that predicts per-hand force commands. A unified force-position whole-body controller then executes the motion and force together, enabling the following capabilities:
\textbf{(b)~\emph{robustness to force variation}}: the robot walks a heavy wheeled gantry forward with both hands;
\textbf{(c)~\emph{bimanual coordination}}: the robot rotates a heavy box (12\,kg) while dragging it across the table without tipping or lifting it;
\textbf{(d)~\emph{sustained contact}}: one hand presses an eraser against a whiteboard and holds a steady normal force through the stroke.
The overlaid curves trace the hand keypoints and the arrows mark the commanded force at its peak; the traces below show the commanded (dashed) and measured (solid) force per hand and axis.}
\vspace{-0.1em}
\label{fig:teaser}
\end{minipage}}
\AddToHook{cmd/@maketitle/after}{\teaserfigure}
\makeatother

\title{\LARGE \bf
Learning Force-Reactive Humanoid Loco-Manipulation
}

\author{Sihang Li$^{1,2,\S}$, Xiaofeng Guo$^{1}$, Jinkun Cao$^{1}$, Charlie Cheng$^{1}$, Ruihan Yang$^{1}$, \\ Chen Feng$^{1,2,\dagger}$, Guanya Shi$^{1,3,\dagger}$, Haozhi Qi$^{1,4,\dagger}$\\[0.3em]
{\normalsize $^{1}$Amazon FAR (Frontier AI \& Robotics)\quad
$^{2}$New York University\quad
$^{3}$Carnegie Mellon University\quad
$^{4}$University of Chicago}%
\thanks{$^{\S}$Work done while interning at Amazon FAR}%
\thanks{$^{\dagger}$Amazon FAR team co-lead}%
}

\begin{document}

\bstctlcite{IEEEtran:BSTcontrol}

\maketitle
\thispagestyle{empty}
\pagestyle{empty}

\begin{abstract}
Many everyday tasks require controlling contact force as well as motion. On a humanoid, this force must often be coordinated across both hands and the whole body, and cannot be specified by kinematic trajectories alone. Therefore, we present \modelname, a framework for learning force-reactive humanoid loco-manipulation. We collect robot-free demonstrations using two force-sensing UMI-style handheld grippers for hand motion and contact wrenches, and wearable trackers for body motion. A force-reactive diffusion policy takes vision, proprioception, and measured wrenches as input, then predicts actions and per-hand force commands. A hybrid compliance whole-body controller with force tracking executes these commands while maintaining balance: an integral of the force error measured by real force/torque (F/T) sensors lets each hand track force accurately along the commanded direction and remain compliant along the others. In simulation, the controller tracks per-hand force commands of varied magnitude and direction, with little loss in whole-body motion accuracy. Across three contact-rich tasks on the robot, the force-reactive policy succeeds in 80\% of rollouts, compared with 26.7\% for a position-only policy trained on the same demonstrations and executed by the same controller. For more details, please visit our website: \url{https://louis-leee.github.io/FLoRe/}.
\end{abstract}

\secv
\section{Introduction}
\secv
\label{sec:introduction}

Humanoid robots offer a general-purpose embodiment for working in environments built for people. Many tasks in these environments, such as pushing a cart, moving a heavy box, and wiping a surface, require the robot to control contact force as well as motion. However, learned humanoid loco-manipulation is commonly formulated as generating and tracking motion references~\cite{BeyondMimic,HOVER,SONIC}. Tracking a demonstrated hand trajectory alone does not reproduce the demonstrated force, which also depends on contact geometry, stiffness, and object motion. Unlike a fixed-base arm, a humanoid must stay balanced under these contact loads, coordinating its hands, torso, and legs while regulating force.


Two difficulties stand in the way. The first is data: imitation learning requires scalable demonstrations that record contact force alongside motion. Existing robot-free interfaces for whole-body demonstration collection record motion but not force~\cite{HuMI,HoMMI}.
The second is execution: the robot must accurately apply the desired force. Current learned humanoid whole-body controllers track motion alone, or treat force as a disturbance to reject or a stiffness to lower~\cite{FALCON,SoftMimic}. Tracking a commanded contact force has been shown only on quadrupedal manipulators, using estimated rather than measured force~\cite{ForceControlLegged,UnifiedForcePos}, which is not accurate enough. A humanoid instead needs to track force at both hands together with whole-body motion while maintaining balance.

We present \modelname, a framework for learning \emph{force-reactive} humanoid loco-manipulation. Its robot-free force-annotated data collection system records contact force with each demonstration; the policy learns to predict this force, and the controller tracks it. The system pairs two UMI-style handheld grippers~\cite{UMI} with a chest tracker and two ankle trackers~\cite{HuMI,BifrostUMI}. Each gripper carries a wrist camera, a pose tracker, and a six-axis F/T sensor~\cite{UMI-FT,ForceMimic}. The applied wrench is thus recorded alongside hand, torso, and foot motion. The robot uses the same gripper--sensor assembly.

For the high-level planner, we propose a force-reactive diffusion policy~\cite{DiffusionPolicy}. It observes wrist images, robot state, and the measured wrench at each hand, then predicts a chunk~\cite{ALOHA-ACT} of action, gripper widths, and per-hand desired force. Force enters as both an input and an output. The input lets the policy react to interactions that images do not reveal. The output is commanded to the controller~\cite{ForceMimic}, rather than used only as an auxiliary target~\cite{TouchDreaming}. This connects the force measured during demonstration to the deployed force.

To track the commanded forces, we develop a whole-body controller with hybrid compliance. It outputs only joint positions, so each hand's force command becomes a shift of its position target. In contact, an integral of the force error measured at the wrist adjusts this shift until the hand exerts the commanded force, whatever the contact stiffness. Along uncommanded axes, the same integral lets the hand yield to external pushes. Each hand thus tracks force along the commanded direction and complies along the others, without switching modes. We train the controller in simulation~\cite{IsaacLab} and deploy it on the robot using wrist F/T measurements.

We deploy \modelname on a Unitree G1 and evaluate three whole-body contact-rich tasks: pushing a gantry hanging another robot, bimanual box rotation, and board wiping. In simulation, the controller realises in-contact force error of about 1\,N per hand while tracking whole-body motion. On the real robot, \modelname succeeds in 80\% of rollouts, compared with 26.7\% for a position-only policy trained on the same demonstrations and executed by the same controller with force commands disabled. Withholding only the force command reduces success to 36.7\%, while removing only the force observation reduces it to 66.7\%. Both contribute to the final performance.
Our key contributions are:
\begin{itemize}[leftmargin=3mm]
  \item A robot-free pipeline for learning force-reactive humanoid loco-manipulation from demonstrations that record both motion and contact wrench.
  \item A learned whole-body controller that tracks per-hand force together with keypoint motion while remaining compliant to external forces and maintaining balance.
  \item Evaluation in simulation and on three contact-rich tasks on a Unitree G1, where force commands and force observations both improve task success over position-only control.
\end{itemize}

\secv
\section{Related Work}
\secv
\label{sec:related_work}

\ssecv
\subsection{Force Sensing for Loco-Manipulation Data Collection and Policy Learning}
\ssecv
\label{sec:rw_data_policy}

Robot-free data collection builds on UMI~\cite{UMI}. Force-aware variants add wrist or fingertip sensing, but deploy on fixed-base arms with analytic impedance or admittance control~\cite{UMI-FT,ForceMimic,OmniUMI}. Humanoid extensions instead collect whole-body motion and execute it with learned trackers, but do not record contact force~\cite{HuMI,BifrostUMI}. WT-UMI is the closest interface for humanoids, although it measures only a scalar normal force at each palm and relies on paired teleoperation data~\cite{WTUMI}. Our interface records a six-axis wrench at each hand together with torso motion, and its demonstrations require no robot-side data.

Measured force has been used to condition visuomotor policies~\cite{FoAR,FACTR}. Other work predicts stiffness, admittance gains, or virtual position targets that are executed by an impedance controller~\cite{VICES,CompACT,AdaptiveCompliancePolicy,UMI-FT}. More closely related methods predict a wrench trajectory together with pose, but execute it on fixed-base arms through analytic force controllers~\cite{ForceMimic,ForcePolicy,FAWAM}. On humanoids, WT-UMI passes a scalar normal-force prediction to an admittance rule~\cite{WTUMI}. UniFP is the closest formulation. It predicts 3-D force commands, but deploys end-effector force control only on a quadrupedal manipulator. Its humanoid interface exposes force only at the base~\cite{UnifiedForcePos}. Our planner instead predicts per-hand pose, gripper, and wrench trajectories from handheld demonstrations, then sends the predicted force to a learned humanoid whole-body controller.

\begin{figure*}[t]
    \centering
    \includegraphics[width=\textwidth]{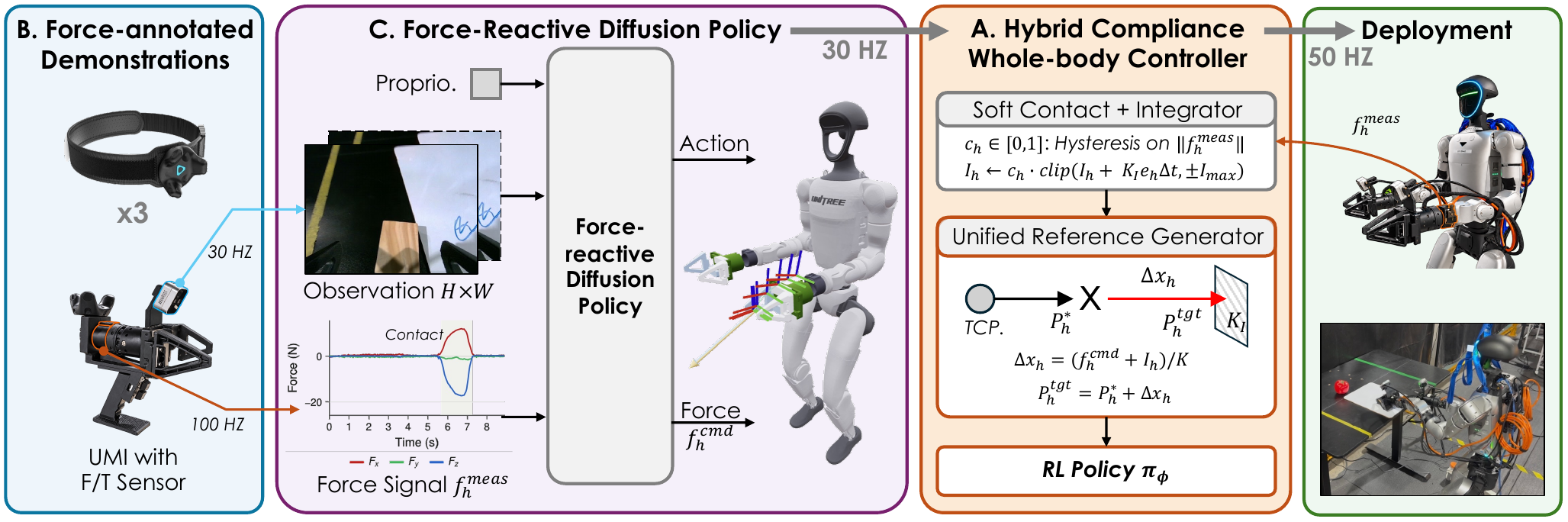}
    \vspace{-10pt}
    \caption{\textbf{Overview of our \modelname framework.} \textbf{(B)}~Two handheld UMI-style grippers with F/T sensors and three additional trackers record images, motion, and contact force without a robot. \textbf{(C)}~A force-reactive diffusion policy takes images, state and measured force as input and outputs keypoint motion and desired force commands at 30\,Hz. \textbf{(A)}~The hybrid compliance whole-body controller with force tracking converts the force command into a shift of the target along the force direction, so that one RL policy tracks motion and force simultaneously while remaining compliant along other directions.}
    \label{fig:overview}
\end{figure*}

\ssecv
\subsection{Force for Low-Level Control}
\ssecv
\label{sec:rw_lowlevel}

Learned legged and humanoid controllers usually either withstand external force or yield to it through commanded stiffness or damping~\cite{FALCON,KinematicsAwareMulti,HEFT,FACET,SoftMimic,CHIP}, but neither approach directly specifies how hard to press. Direct force tracking has mainly been studied on quadrupedal manipulators, as a separate control mode or through a target-shifting spring law, both using force estimated from proprioception~\cite{ForceControlLegged,UnifiedForcePos}. WT-UMI regulates a scalar palm force on a humanoid, but through an analytic admittance rule and inverse kinematics above a separate locomotion controller~\cite{WTUMI}. In contrast, our hybrid compliance whole-body controller with force tracking integrates the force error measured at each wrist. One learned policy tracks position in free space and force along the commanded direction, while remaining compliant along the other directions and maintaining balance.

\secv
\section{Method}
\secv
\label{sec:method}

In \modelname (Fig.~\ref{fig:overview}), a hybrid compliance whole-body controller with force tracking executes motion references and per-hand force commands (Sec.~\ref{sec:method_controller}); a robot-free force-annotated data collection system records vision, motion, and contact wrenches with handheld devices (Sec.~\ref{sec:method_data}); and a force-reactive diffusion policy, trained on these demonstrations, predicts keypoint motion and force commands for the controller (Sec.~\ref{sec:method_policy}).

\noindentp{Notation.} The robot is a humanoid with a parallel-jaw gripper on each hand. The controller tracks keypoints $k\in\keypts$ (both hand tool center point (TCP), the torso, and both feet) with poses $\pose_k=(\pos_k,\rot_k)$ expressed in the pelvis frame. For each hand $h\in\hands$, $\fcmd_h$ is the force the hand should exert on the environment, $\fmeas_h$ the reaction force measured by its wrist F/T sensor, and $\fext_h$ the true contact force available only in simulation; all are expressed in the pelvis frame and act at the TCP, so that perfect tracking gives $\fmeas_h=-\fcmd_h$. The high-level policy predicts chunks of $\horizon$ steps; the low-level controller runs at a higher rate and sends joint-position targets to joint PD controllers. Hardware, rates, and all numerical settings are given in Sec.~\ref{sec:exp_impl}.

\ssecv
\subsection{Hybrid Compliance Whole-body Controller with Force Tracking Capability}
\ssecv
\label{sec:method_controller}
The controller is a single learned policy that tracks whole-body motion together with a commanded force at each hand. We describe its control law, the contact model and simulated sensor it is trained with, and its training objective.

\begin{figure}[t]
    \centering
    \includegraphics[width=0.8\linewidth]{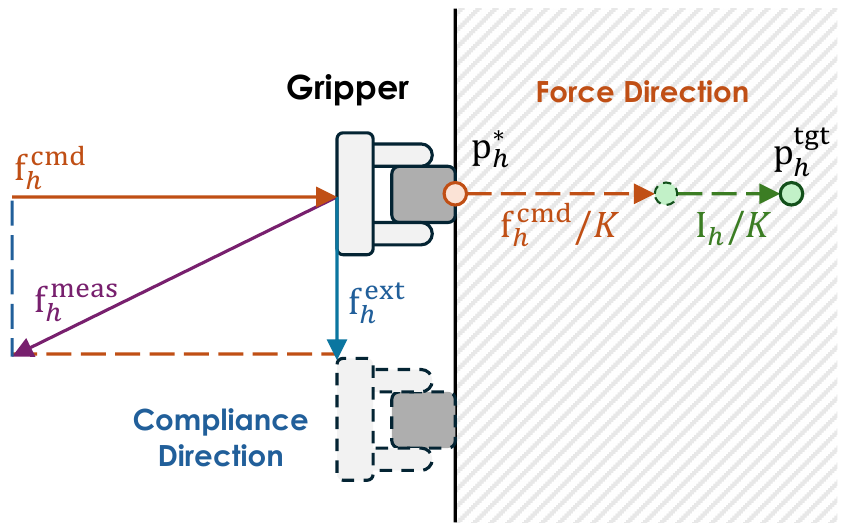}
    \vspace{-10pt}
    \caption{\textbf{Control law of the hybrid compliance controller.} Along the force direction, the target $\pos^{\mathrm{tgt}}_h$ is shifted from the motion reference $\pos^{*}_h$ into the surface by the force command and the force feedback, until the measured force $\fmeas_h$ balances the command. Along the compliance direction, the gripper yields to $\fext_h$ (dashed gripper).}
    \vspace{-5pt}
    \label{fig:controller}
\end{figure}

\noindentp{Control law and unified reference generator.} The controller outputs only joint positions and is rewarded only on keypoint poses, so a force command must become a position target. For each hand $h$, a unified reference generator combines the motion reference $\pos^{*}_h$ and the force command $\fcmd_h$ into a single target (see Fig.~\ref{fig:controller}):
\begin{equation}
\begin{aligned}
\pos^{\mathrm{tgt}}_h &= \underbrace{\pos^{*}_h}_{\text{motion reference}} + \underbrace{\frac{\fcmd_h}{K}}_{\text{force command}} + \underbrace{\frac{\integ_h}{K}}_{\text{force feedback}},\\
\integ_h &\leftarrow \cphase_h\,\mathrm{clip}\big(\integ_h + K_I\underbrace{(\fcmd_h+\fmeas_h)}_{\text{force error}}\Delta t,\,\pm I_{\max}\big),
\end{aligned}
\label{eq:target_shift}
\end{equation}
with nominal stiffness $K$, integral gain $K_I$, bound $I_{\max}$ and time step $\Delta t$; $\dx_h=(\fcmd_h+\integ_h)/K$ is the total shift. The command moves the target $\fcmd_h/K$ along the force direction; where the target lies inside the surface, the hand cannot follow it and presses instead, and tracking means $\fmeas_h=-\fcmd_h$. How hard a given shift presses depends on the contact stiffness, which the robot does not know. The integral of the force error therefore moves the target further in, or back, until the measured force balances the command, independent of $K$ and $K_{\mathrm{env}}$; unlike the spring law of UniFP~\cite{UnifiedForcePos}, which shifts the target in proportion to an estimated contact force, the shift is thus closed on the measured force. The soft contact phase $\cphase_h\in[0,1]$, computed from the measured force magnitude with onset and release hysteresis thresholds and a first-order low-pass filter, lets the integral build up only in contact and clears it on release, while the clip bounds windup. It requires no privileged contact information, and also enters the observation and weights the force-related rewards. $\pos^{\mathrm{tgt}}_h$ replaces the hand target in the tracking reward, and $\integ_h$ and $\dx_h$ enter the observation; the joint command is untouched, so the policy remains the sole executor.

The law acts on every axis of each hand, and a measured force need not lie along the command. Dropping the hand index, let a force be commanded along $x$ and none along $z$. A push along $z$ is then measured too, $f^{\mathrm{meas}}_z=f^{\mathrm{ext}}_z$:
\begin{equation*}
\begin{aligned}
x&:\ f^{\mathrm{cmd}}_x\neq0 \Rightarrow f^{\mathrm{meas}}_x\to-f^{\mathrm{cmd}}_x \quad\text{(force tracking)},\\
z&:\ f^{\mathrm{cmd}}_z=0 \Rightarrow \dot{I}_z=K_I f^{\mathrm{meas}}_z=K_I f^{\mathrm{ext}}_z \quad\text{(compliance)}.
\end{aligned}
\end{equation*}
Along $z$ the integral grows with the push and moves the target with it at a rate $K_I f^{\mathrm{ext}}_z/K$, so the hand gives way instead of resisting, for as long as the push lasts and within the bound $I_{\max}$. Each hand thus tracks force along the commanded direction and remains compliant along the others, with no switching between them.

\noindentp{Contact model and simulated sensor.} In training each hand faces a virtual compliant plane with outward normal $\mathbf n_h$ and randomized stiffness $K_{\mathrm{env}}$ that reacts only to penetration,
\begin{equation}
\fext_h = K_{\mathrm{env}}\,\max\!\big(0,\ \mathbf{n}_h^{\!\top}(\mathbf{x}^{\mathrm{wall}}_h-\pos_h)\big)\,\mathbf{n}_h,
\end{equation}
applied at the TCP, with penetration capped at $d_{\max}$ and the force capped as well. The policy observes a simulated F/T reading instead of $\fext_h$,
\begin{equation}
\fmeas_h = \fext_h + m_d\,\ddot{\pos}_h^{\mathrm{wrist}} + \mathbf{b}_h + \boldsymbol{\epsilon}_h ,
\end{equation}
with randomized distal mass $m_d$, per-episode bias $\mathbf b_h$ and white noise $\boldsymbol\epsilon_h$. At each reset the plane is placed at a random distance beyond the hand's reset position with a random orientation, and $\fcmd_h$ follows trapezoidal ramp--hold--release pulses with random peaks, clamped to what the plane can deliver, $K_{\mathrm{env}}d_{\max}$. The motion reference is the unmodified clip; where it conflicts with the force, the shifted target of \eqref{eq:target_shift} wins. The plane is frictionless, so training commands are purely normal; on the robot the same per-axis law applies along tangential axes as above, although tangential loads are never seen in training.

\begin{figure}[t]
    \centering
    \includegraphics[width=1\linewidth]{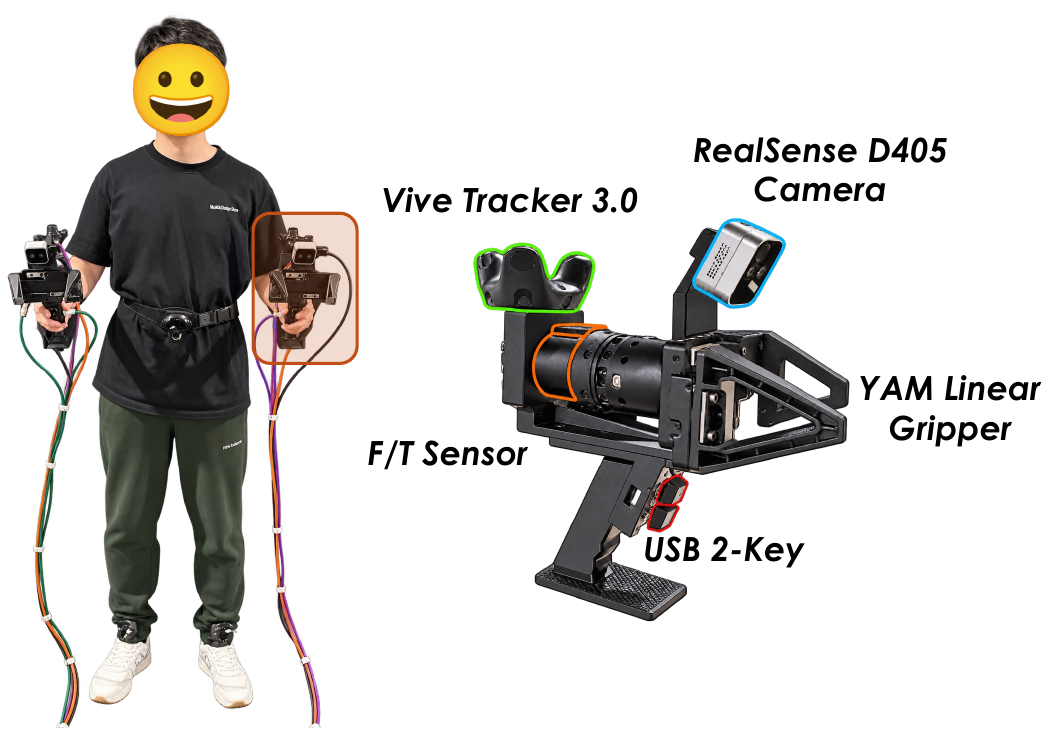}
    \vspace{-20pt}
    \caption{\textbf{Robot-free data-collection interface.} Left: the demonstrator holds a handheld device in each hand and wears trackers on the chest and both ankles. Right: each device carries a VIVE tracker, a RealSense D405 wrist camera, and an F/T sensor mounted between the handle and gripper.}
    \vspace{-15pt}
    \label{fig:device}
\end{figure}

\noindentp{Training objective.} The reward extends the motion-tracking reward of~\cite{BeyondMimic}, $r_{\mathrm{track}}$, with terms for the hands:
\begin{equation}
\begin{aligned}
r = r_{\mathrm{track}} &+ \,r_{\mathrm{pos}} + \,r_{\mathrm{tcp}} + \,r_{\mathrm{orn}} + \,r_{\mathrm{force}} \\
&+ \,r_{\mathrm{contact}} + \,r_{\mathrm{approach}} - \,r_{\mathrm{rate}} .
\end{aligned}
\label{eq:reward}
\end{equation}
All tracking terms use the shifted target of \eqref{eq:target_shift}: $r_{\mathrm{pos}}$, $r_{\mathrm{tcp}}$ and $r_{\mathrm{orn}}$ are exponential kernels on the hand-position, TCP-position and hand-orientation errors, the latter two constraining the wrist-to-TCP offset. Weighted by $\cphase_h$ while a force is commanded, $r_{\mathrm{force}}$ scores the force error against the clean simulated force and $r_{\mathrm{contact}}$ rewards contact below the force cap; weighted by $1-\cphase_h$, $r_{\mathrm{approach}}=\exp(-d_h/d_0)$ pulls the hand toward the plane, with $d_h$ its distance and $d_0$ a length scale. $r_{\mathrm{rate}}$ penalizes changes in contact force between steps and softens impacts.

\ssecv
\subsection{Force-annotated Robot-free Data Collection System}
\ssecv
\label{sec:method_data}

\noindentp{Device.} Each handheld UMI-style gripper~\cite{UMI} carries an F/T sensor\footnote{\url{https://botasys.com/force-torque-sensors/rokubi/}} between handle and gripper, a wrist camera, a pose tracker, and a jaw-width encoder; the demonstrator holds one in each hand and wears trackers on the chest and both ankles (Fig.~\ref{fig:device}). The robot carries the same gripper--sensor assembly on each wrist, so sensor frame, sensor-to-TCP transform and gripper geometry are identical in collection and deployment. The wrench is kept in the sensor frame, gravity-compensated with a per-device payload and tared at the start of each recording; on the robot, inertial loads are also compensated.

\noindentp{Streams and synchronization.} Images, tracker poses, wrenches, and gripper widths arrive at different rates and are stamped on arrival with one host clock; the labels below are built from the raw streams, not the camera grid. Missing wrench readings and tare transients are masked rather than zero-filled, since a tared sensor also reads zero in free space.

\ssecv
\subsection{Force-Reactive High-level Policy Learning}
\ssecv
\label{sec:method_policy}
The high-level policy $\pihl$ is a diffusion policy~\cite{DiffusionPolicy} that maps the current observation to a chunk of $\horizon$ future actions~\cite{ALOHA-ACT}. We augment both its inputs and outputs with per-hand contact wrenches.

\noindentp{Observation.} The observation $\obs_t$ contains the wrist images of both hands, $\img^L_t$ and $\img^R_t$, the measured wrenches $\{\wmeas_{h,t}\}_{h\in\hands}$, and a low-dimensional state:
\begin{equation}
\begin{aligned}
\state_t=\big[&(\pos_{H_L}-\pos_{H_R})_{xy},\ (\pos_{H_L}-\pos_T)_{xy},\ (\pos_{H_R}-\pos_T)_{xy},\\
&\mathrm{rpy}(\rot_{H_L}),\ \mathrm{rpy}(\rot_{H_R}),\ g_{H_L},\ g_{H_R}\big].
\end{aligned}
\end{equation}
The state contains pairwise horizontal position differences among the hands and torso, hand orientations in the pelvis frame, and gripper widths $g_h$. Here, $(\cdot)_{xy}$ selects the horizontal components, and $\mathrm{rpy}(\rot)$ returns the extrinsic-XYZ Euler angles (roll, pitch, yaw) of $\rot$ in radians. Using position differences makes the state invariant to absolute translation. We omit absolute poses and joint angles because they differ between the completed demonstration motion and the robot. 

\noindentp{Action.} Each step $j$ of the chunk stacks the pose, gripper width and force command $\fcmd_h$ of each hand, together with the poses of the torso and both feet:
\begin{equation}
\begin{aligned}
\act_{t+j}&=[\{\relpose_h,g_h,\fcmd_h\}_{h\in\hands},\{\relpose_k\}_{k\in\keypts\setminus\hands}],\\
\relpose_k&=\pose_{k,t}^{-1}\,\pose_{k,t+j}.
\end{aligned}
\end{equation}
Poses are \emph{anchor-relative}, in each keypoint's own frame at the observation time, so the chunk is invariant to the world frame and to where the demonstrator stood~\cite{HuMI,BifrostUMI}; rotations use the continuous rotation representation of~\cite{Rot6D}.

\noindentp{Model and training.} Each image is encoded by a CLIP-initialized vision transformer~\cite{CLIP}, the state by a linear layer, and each hand's wrench history by a small causal convolutional encoder, in which a learned embedding replaces masked windows. These condition a temporal U-Net~\cite{UNet} denoiser through FiLM~\cite{FiLM}, trained with the DDPM objective~\cite{DDPM} and sampled with DDIM~\cite{DDIM}. Actions are quantile-normalized per column and the loss is a masked Mean Squared Error (MSE) Loss.

\noindentp{Closing the loop.} At each policy step the chunk is decoded into absolute pelvis-frame keypoint targets, re-based onto the currently commanded keypoints so that a re-plan never jumps, and streamed to the controller either sequentially (execute $N$ steps, then re-plan) or asynchronously with temporal ensembling~\cite{ALOHA-ACT,HATO}. The predicted force of each hand becomes its $\fcmd_h$ (Sec.~\ref{sec:method_controller}).

\secv
\section{Experiments}
\secv
\label{sec:experiments}

We evaluate whether a humanoid can learn force-reactive loco-manipulation from robot-free demonstrations. Experiments in simulation and on the robot test four capabilities:
\begin{itemize}[leftmargin=3mm]
  \item \textbf{Whole-body force tracking}: tracking per-hand force commands and whole-body motion with one learned controller.
  \item \textbf{Sustained contact}: making contact and holding an appropriate force while the hands move with or along the object.
  \item \textbf{Bimanual and whole-body coordination}: applying force with both hands while the torso and legs balance and walk.
  \item \textbf{Robustness to force variation}: completing tasks when resistance changes within or across episodes but is not visible in the images, e.g., a heavier load or box mass.
\end{itemize}

We first compare our hybrid compliance whole-body controller with force tracking (Sec.~\ref{sec:method_controller}) to three baselines:
\begin{itemize}[leftmargin=3mm]
  \item \bl{SONIC}~\cite{SONIC}: a large-scale whole-body motion tracker, run from its released checkpoint. It tracks motion references and accepts no force command.
  \item \bl{SONIC + admittance}: SONIC followed by an analytic force servo that adds $\alpha\,\mathbf J_h^{\!\top}(\fcmd_h+\fmeas_h)$ to its arm joint targets, a correction that SONIC does not observe.
  \item \bl{\modelname w/o Force}: our controller trained with the force channel removed ($\fcmd\equiv0$, no $\integ_h,\dx_h,\cphase_h$ observed).
\end{itemize}

The second comparison isolates how the high-level policy uses force. \modelname observes the measured wrench (\emph{F-in}), learns to predict the demonstrated force (\emph{F-out}), and sends this prediction to the controller (\emph{F-cmd}). \bl{w/o F-cmd} and \bl{w/o F-in} remove one pathway each, and \bl{DP} all three:
\begin{itemize}[leftmargin=3mm]
  \item \bl{w/o F-cmd}: the \modelname policy with its predicted force withheld ($\fcmd\equiv0$), which turns off the force channel ($\cphase_h,\integ_h,\dx_h\equiv0$) so that the controller tracks pose only.
  \item \bl{w/o F-in}: the \modelname policy trained without the measured wrench in its observation. It still commands force but cannot sense whether contact is made or lost.
  \item \bl{DP}: the position-only diffusion policy of UMI-style humanoid systems~\cite{HuMI,BifrostUMI}, trained with the same architecture, data and schedule as \modelname but without force input or output, and executed by our controller with $\fcmd\equiv0$.
\end{itemize}

We train one policy per task and evaluate each in 10 rollouts, with test conditions paired across policies. We report the success rate, compared using Fisher's exact test, and the offline validation error of the predicted pose, force and torque on held-out demonstrations.

\begin{figure}[t]
    \centering
    \includegraphics[width=1\linewidth]{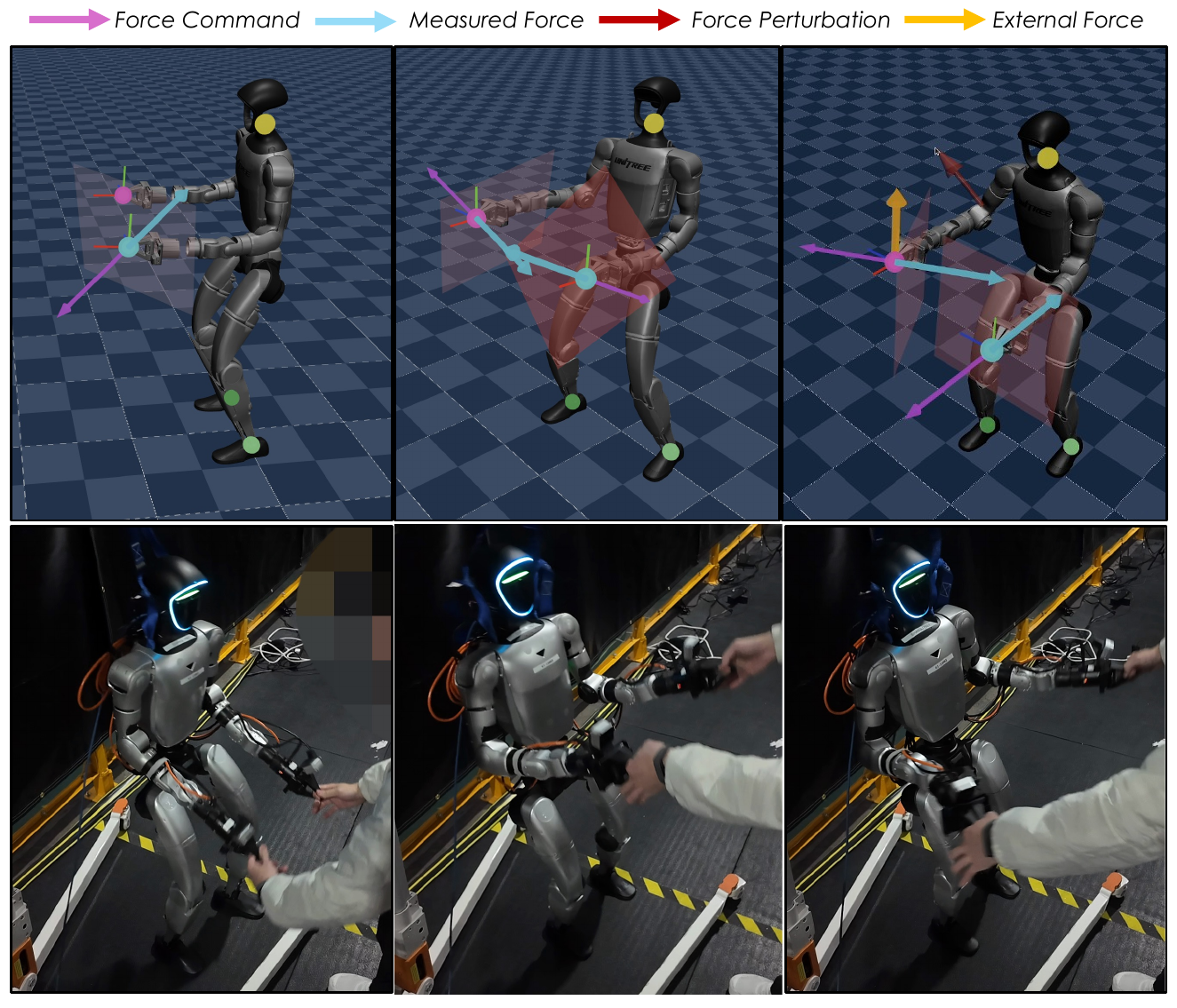}
    \vspace{-18pt}
    \caption{\textbf{Hybrid compliance whole-body controller with force tracking.} Top: in simulation, each hand presses a randomly placed virtual plane while the robot tracks the reference motion and maintains balance. The arrows show commanded force (pink), measured reaction force (blue), external force (yellow), and force perturbation (red). Bottom: on the real robot, a person pulls both hands, which yield instead of resisting.}
    \label{fig:sim_training}
\end{figure}

\ssecv
\subsection{Implementation Details}
\ssecv
\label{sec:exp_impl}

\begin{table*}[tb]
\centering
\caption{\label{tab:ll_tracking} Sim-to-sim evaluation of motion and force tracking for our unified force--position controller and the baselines. Motion tracking is evaluated on two held-out motion sets. $^\dagger$Averaged over the left and right hands.}
\footnotesize
\renewcommand{\arraystretch}{1.15}
\setlength{\tabcolsep}{2.5pt}
\begin{tabular*}{\textwidth}{@{\extracolsep{\fill}}lccccccc@{}}
\toprule
 & & \multicolumn{3}{c}{\textbf{Motion Tracking}} & \multicolumn{3}{c}{\textbf{Force Tracking}$^\dagger$} \\
\cmidrule(lr){3-5} \cmidrule(lr){6-8}
\textbf{Controller} & $\fcmd$ & Global Root (m) $\downarrow$ & Wrist Pos (mm) $\downarrow$ & Wrist Ori (rad) $\downarrow$ & $a_1$ $\rightarrow$1 & $|e_F|$ (N) $\downarrow$ & $F{=}0$ floor (N) $\downarrow$ \\
\midrule
\multicolumn{8}{@{}l}{\textit{ROOT-90}~\cite{sim2real2026}} \\
\midrule
\bl{SONIC}~\cite{SONIC}                    & \xmark & \textbf{0.317} & 47.16 & 0.2787 & N/A         & N/A       & N/A \\
\bl{SONIC + admittance}~\cite{SONIC}       & \cmark & \textbf{0.317} & 47.16 & 0.2787 & 0.70 & 7.89 & 2.62 \\
\bl{\modelname w/o Force}                        & \xmark & 0.342 & \textbf{36.75} & 0.2239 & N/A         & N/A       & N/A \\
\modelname(\textbf{\ours})                 & \cmark & 0.374 & 44.70 & \textbf{0.1176} & \textbf{1.06} & \textbf{1.05} & \textbf{1.50} \\
\midrule
\multicolumn{8}{@{}l}{\textit{VLK-Pick-100}~\cite{wang2026vlk}} \\
\midrule
\bl{SONIC}                                 & \xmark & 0.072 & \textbf{21.45} & 0.2658 & N/A         & N/A       & N/A \\
\bl{SONIC + admittance}~\cite{SONIC}       & \cmark & 0.072 & \textbf{21.45} & 0.2658 & 0.67 & 7.87 & 2.62 \\
\bl{\modelname w/o Force}                        & \xmark & \textbf{0.071} & 23.54 & 0.2141 & N/A         & N/A       & N/A \\
\modelname(\textbf{\ours})                 & \cmark & 0.084 & 23.16 & \textbf{0.1165} & \textbf{1.03} & \textbf{1.00} & \textbf{1.25} \\
\bottomrule
\end{tabular*}

\end{table*}

\noindentp{Hybrid Compliance Whole-body Controller.} The reference generator uses $K=200$\,N/m, $K_I=0.5\,\mathrm{s}^{-1}$ and $I_{\max}=40$\,N; the soft contact phase uses hysteresis thresholds of 2.0\,N for onset and 1.5\,N for release and a 50\,ms low-pass time constant. We sample $K_{\mathrm{env}}\sim\mathrm{LogUniform}(100,800)$\,N/m and cap penetration at $d_{\max}=8$\,cm and contact force at 35\,N. Command pulses peak at up to 25\,N. The reward kernels have widths of 0.1\,m (TCP position), 0.25\,rad (hand orientation) and 25\,N (force). $r_{\mathrm{contact}}$ rewards contact below 35\,N.

\noindentp{Policy.} The two $224\times224$ wrist images are encoded by a CLIP-initialized ViT-B/16; the state has 14 dimensions, and rotations use the first two rows of the rotation matrix. The 1-D U-Net denoiser is sampled with 16 DDIM steps, the image encoder is trained with a $10\times$ lower learning rate, and each policy is trained for $10^5$ steps with $\horizon=50$ (1.67\,s).

\ssecv
\subsection{Hybrid Compliance Whole-body Low-level Controller}
\ssecv
\label{sec:exp_ll}

\noindentp{Simulation protocol.} Following~\cite{BeyondMimic}, we train in Isaac Lab~\cite{IsaacLab} and evaluate sim-to-sim in MuJoCo (Fig.~\ref{fig:sim_training}). Motion/Force tracking is measured with $\fcmd=0$ on two held-out sets of retargeted clips: ROOT-90~\cite{sim2real2026} and VLK-Pick-100~\cite{wang2026vlk}. For force tracking, the dose-response test applies forward commands of 5, 10, 15, 20, and 25\,N to each hand in turn, with $F{=}0$ as the baseline; we also test 10\,N commands along $\pm x$, $\pm y$, and $\pm z$, and simultaneous commands to both hands. Statistics are computed over clips in which both paired conditions complete without a fall.

\noindentp{Metrics.}  For motion tracking, we report global root error and wrist position and orientation errors~\cite{MaskedMimic}. For force tracking, the gain $a_1$ is the slope of a least-squares fit, with free intercept, of the realized hold-phase force against the true contact force $\fext$. The in-contact error $\norm{e_F}_1=\norm{\fcmd_h+\fmeas_h}_1$ is averaged over the hold phase of each pulse. The $F{=}0$ floor is the force read back when nothing is commanded; it bounds the smallest verifiable command.

\noindentp{Adding a force channel costs little motion tracking.}
Table~\ref{tab:ll_tracking} compares \modelname with \bl{\modelname w/o Force}, \bl{SONIC}~\cite{SONIC}, and \bl{SONIC + admittance}, all evaluated with $\fcmd=0$ on two held-out motion sets. Relative to \bl{\modelname w/o Force}, \modelname adds 1.3--3.2\,cm of global root error and at most 8.0\,mm of wrist position error, while its wrist orientation error roughly halves; it also tracks orientation better than \bl{SONIC}, at a higher root error. When a force is commanded, the hand target is shifted by design ($\fcmd/K$ alone is 12.5\,cm at 25\,N), and the hand departs from the motion reference only as far as needed to realize the force.

\noindentp{Hybrid compliance controller tracks commanded force accurately.}
Neither \bl{SONIC} nor \bl{\modelname w/o Force} accepts a force command. \bl{SONIC + admittance} does, but its servo acts outside the balance policy and realizes only 0.67--0.70 of the commanded force, with an in-contact error of about 7.9\,N. \modelname tracks 5--25\,N commands with a gain of 1.03--1.06 and about 1\,N of in-contact error per hand; lateral and downward commands reach a lower gain and leak partly into the other axes. With nothing commanded the sensor still reads 1.25--1.50\,N, so at deployment we zero commands below 1.5\,N and cap them at 25\,N. Integrating the measured force error removes the dependence on the unknown contact stiffness: \bl{Spring-Law}, the target-shift law of~\cite{UnifiedForcePos} without its force-feedback term, cannot detect contact, and its force scales with that stiffness relative to the assumed $K$. The same integral makes the hand compliant: while it tracks a force, a push along an uncommanded axis is read by the wrist sensor and builds up the integral, which moves the target with it, so the hand gives way instead of resisting, with no mode switch (real robot, Fig.~\ref{fig:sim_training} bottom).

\begin{figure*}[t]
    \centering
    \includegraphics[width=1\linewidth]{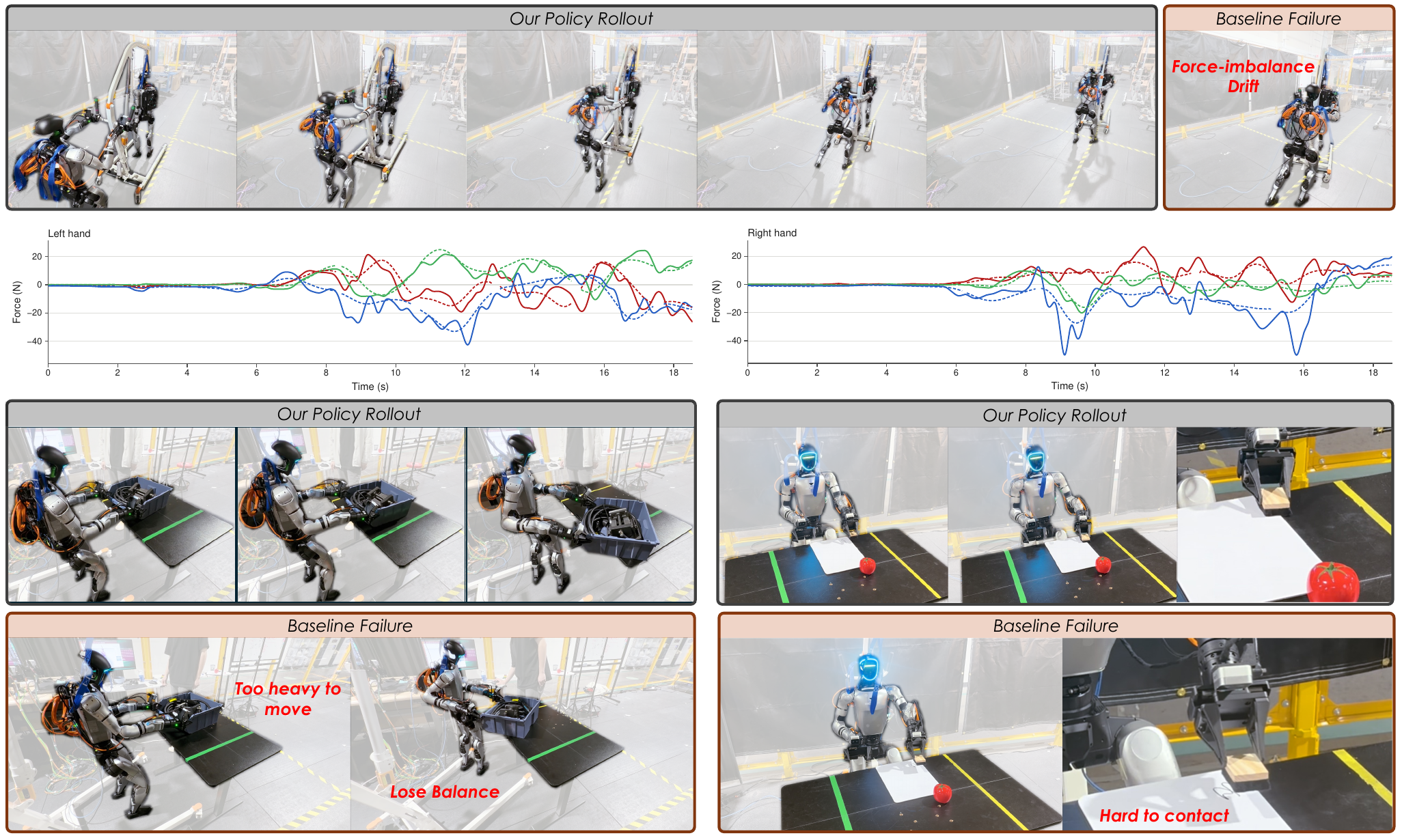}
    \vspace{-18pt}
    \caption{\textbf{Rollouts on the robot and failure modes of the position-only baseline.}
      Top: \ours walks the gantry forward from left to right. The baseline either fails to move the gantry or pushes unequally, causing it to drift from its heading. The traces show commanded (dashed) and measured (solid) force for each hand and axis.
      Bottom left: \ours drags and rotates the box. The baseline either cannot move it or leans into it and loses balance.
      Bottom right: \ours erases the strip under steady pressure. The baseline hovers above the board and never makes contact.}
    \label{fig:visualization}
\end{figure*}

\begin{table*}[tb]
\centering
\caption{\label{tab:hl_main} Success rate and offline validation error of the predicted pose, force and torque for \ours and its force-pathway ablations, trained on the same demonstrations and executed by the same unified controller.}
\footnotesize
\renewcommand{\arraystretch}{1.15}
\setlength{\tabcolsep}{3.5pt}
\begin{tabular*}{\textwidth}{@{\extracolsep{\fill}}lcccccccccc@{}}
\toprule
 & \multicolumn{3}{c}{\textbf{Force Switches}} & \multicolumn{4}{c}{\textbf{Success Rate (\%)} $\uparrow$} & \multicolumn{3}{c}{\textbf{Evaluation Error (Val)} $\downarrow$} \\
\cmidrule(lr){2-4} \cmidrule(lr){5-8} \cmidrule(lr){9-11}
\textbf{Policy} & F-in & F-out & F-cmd & \task{push\_gantry} & \task{rotate\_box} & \task{wipe\_board} & Mean & Pose (cm) & Force (N) & Torque (N\,m) \\
\midrule
\textbf{\ours}   & \cmark & \cmark & \cmark & 7/10 & 9/10 & 8/10 & 80\% & 1.18 & 1.29 & 0.148 \\
\midrule
\bl{w/o F-cmd$^\dagger$}   & \cmark & \cmark & \xmark & 3/10 & 4/10 & 4/10  & 36.7\% & 1.18 & 1.29 & 0.148 \\
\bl{w/o F-in}    & \xmark & \cmark & \cmark & 6/10 & 7/10 & 7/10 & 66.7\% & 1.78 & 2.51 & 0.277 \\
\bl{DP}          & \xmark & \xmark & \xmark & 2/10 & 3/10 & 3/10 & 26.7\% & 2.23 & N/A & N/A \\
\bottomrule
\end{tabular*}
\par\vspace{1pt}
{\scriptsize\raggedright $^\dagger$Predicted and supervised but withheld from the controller ($\fcmd=0$): auxiliary supervision only. Offline, \bl{w/o F-cmd} is the same network as \ours. }
\end{table*}

\noindent\textbf{Real-world force tracking.}
We verify force tracking on the robot against an external reference: a rigidly mounted digital force gauge that the hand presses along each of $\pm x$, $\pm y$, and $\pm z$ while the humanoid stands freely. The command follows a staircase of 4, 6, 8, and 10\,N, repeated across directions and hands. Along the $x$ axis, the measured steady-state force reaches approximately $91\%$ of the commanded magnitude, with a mean absolute steady-state error of about 1.1\,N. The response enters a $\pm1$\,N band in 1.8\,s on average, while the steady-state fluctuation remains approximately $\pm0.3$--$0.5$\,N. The controller consistently recovers the commanded contact force without loss of stability.
\ssecv
\subsection{Push Gantry}
\ssecv
\noindentp{Task.} As shown in Fig.~\ref{fig:teaser}b, the robot places both hands on a heavy wheeled gantry, from which a second humanoid hangs, and walks it forward to a target distance. The demonstrated hand motion does not specify how hard to push, yet the gantry does not move until the push overcomes its static friction. This is the only task recorded with ankle trackers, so the policy predicts all five keypoints. A rollout succeeds if the gantry reaches the target distance without a stop.

\noindentp{Capabilities.}
\underline{Bimanual and whole-body coordination:} the legs and torso produce the push while the two hands balance their forces; otherwise the gantry drifts from its heading.
\underline{Sustained contact:} both hands must keep contact as the gantry accelerates and the robot steps.
\underline{Robustness to force variation:} the hanging humanoid swings as the gantry rolls, so the resistance changes over time.


\noindentp{Test scenarios.} We run 10 rollouts per policy over two gantry loads and five initial poses that vary in distance and heading.

\noindentp{Performance.} \ours reaches the target in 7/10 rollouts (Table~\ref{tab:hl_main}). It fails when a strong push coincides with a step. \bl{DP} succeeds in 2/10 rollouts; in the others the gantry either does not move or drifts from its heading as the two hands push unequally (Fig.~\ref{fig:visualization}, top). \bl{w/o F-cmd} succeeds in 3/10 and fails similarly, as it senses the resistance but cannot command force. \bl{w/o F-in} succeeds in 6/10. Without force feedback, it stalls against the heavier load and keeps walking after losing contact with the lighter one. Offline, \ours has a pose error of 0.82\,cm and a force error of 0.77\,N per hand.

\ssecv
\subsection{Rotate Box}
\ssecv
\noindentp{Task.} As shown in Fig.~\ref{fig:teaser}c, the robot pinches the rim of a box with both hands. It rotates the box by about $50^\circ$, drags it about 0.5\,m across the table, and releases it. The box remains on the table, so the measured wrench reflects table friction, box inertia, and the internal squeeze between the hands. A rollout succeeds if the box rotates by more than $40^\circ$ and remains upright after release.

\noindentp{Capabilities.}
\underline{Bimanual coordination:} the hands share an internal force and rotate about a common axis. In the demonstrations, the right hand applies about 1.4$\times$ the force of the left. If one hand leads the other, the box tips.
\underline{Sustained contact:} the pinch must hold through the drag and the yaw without slipping or lifting the box.
\underline{Robustness to force variation:} friction changes with the box mass and when the box begins to slide.


\noindentp{Test scenarios.} We run 10 rollouts per policy over two box masses and five initial yaw angles.

\noindentp{Performance.} \ours succeeds in 9/10 rollouts and has an offline pose error of 0.70\,cm. \bl{DP} succeeds in 3/10 and exhibits two failure modes (Fig.~\ref{fig:visualization}, bottom left). It either fails to drag the heavy box or leans into it and loses balance. \bl{w/o F-cmd} succeeds in 4/10 and fails similarly because its predicted squeeze is not executed. \bl{w/o F-in} succeeds in 7/10 but cannot react when the box begins to slide.

\ssecv
\subsection{Wipe Board}
\ssecv
\noindentp{Task.} As shown in Fig.~\ref{fig:teaser}d, the robot presses an eraser against a whiteboard and wipes a marked strip. Success depends on normal force, which is not visible in the images. A rollout succeeds if the strip is erased without a stop.

\noindentp{Capabilities.}
\underline{Sustained contact:} the robot must maintain normal force during each stroke and re-establish it after losing contact. Images alone do not distinguish hovering from pressing.
\underline{Robustness to force variation:} we displace the board by $\pm3$\,cm along its normal. The demonstrated trajectory then produces either no contact or excessive force if tracked directly.


\noindentp{Test scenarios.} We run 10 rollouts per policy over two board offsets ($\pm3$\,cm along the normal) and two surfaces.

\noindentp{Performance.} \ours erases the strip in 8/10 rollouts. \bl{DP} succeeds in 3/10. It usually hovers above the board (Fig.~\ref{fig:visualization}, bottom right). This directly shows that observed motion can be valid supervision in free space but not in contact~\cite{BRIDGE}. \bl{w/o F-cmd} succeeds in 4/10. It detects contact but cannot increase a force that is too low. \bl{w/o F-in} succeeds in 7/10 but continues wiping in free space when the board is farther away. This task also has the largest offline errors: 2.02\,cm for pose and 1.59\,N for force.
\ssecv
\subsection{Findings Summary}
\ssecv
\label{sec:exp_findings}
\noindentp{F1: The force channel costs little motion tracking.} Relative to \bl{\modelname w/o Force}, \modelname adds 1.3--3.2\,cm of root error and at most 8.0\,mm of wrist position error while halving wrist orientation error (Table~\ref{tab:ll_tracking}); in return it tracks commanded force with a gain of 1.03--1.06 and about 1\,N of in-contact error. The 1.25--1.50\,N $F{=}0$ floor bounds the smallest verifiable command, and lateral and downward commands remain weaker.

\noindentp{F2: Force-reactive policies resolve failures that motion alone cannot distinguish.} With the same demonstrations and controller, \bl{DP} succeeds in 26.7\% of rollouts and \modelname in 80\% (Table~\ref{tab:hl_main}). The position-only policy stalls when a reached pose does not move the object, hovers or presses too hard when the board is displaced, and loses balance against the heavy box (Fig.~\ref{fig:visualization}). Force also composes with coordination: both hands hold the internal force on Rotate Box, and the push is sustained while walking on Push Gantry.

\noindentp{F3: Force commands give the larger, significant gain; force observations show a smaller trend.} Withholding the predicted force from the controller reduces success from 80\% (24/30) to 36.7\% (11/30; Fisher's exact test, $p=0.001$): the policy senses resistance but can respond only through pose. Removing the force observation reduces success to 66.7\%.

\secv
\section{Conclusion}
\secv
\label{sec:conclusion}

We presented \modelname, a framework for learning force-reactive humanoid loco-manipulation from robot-free demonstrations that record both motion and contact wrench. Our hybrid compliance whole-body controller tracks the learned force commands, allowing the humanoid to sustain contact, coordinate both hands, and apply force while walking and balancing. On three tasks, the force-reactive policy succeeds in 80\% of rollouts, compared with 26.7\% for a position-only policy. By making force a shared interface between demonstrations, policies, and whole-body control, \modelname extends robot-free imitation learning to contact-rich skills that motion alone cannot specify.

\bibliographystyle{IEEEtran}
\bibliography{references,bibcontrol}

\end{document}